\documentclass[10pt,twocolumn,letterpaper]{article}

\usepackage{cvpr}
\usepackage{times}
\usepackage{epsfig}
\usepackage{graphicx}
\usepackage{amsmath}
\usepackage{amssymb}
\usepackage{lipsum}
\usepackage{siunitx}
\usepackage{gensymb}
\usepackage{tabularx}
\usepackage{multirow}
\usepackage{booktabs}
\usepackage{colortbl}
\usepackage{xfrac}
\usepackage{tikz}
\usetikzlibrary{calc,shapes,arrows}

\usepackage{xcolor,colortbl}
\definecolor{bestcolor}{RGB}{180,255,130}
\newcommand{\bestr}[1]{\underline{#1}}
\newcommand{\bestc}[1]{\textbf{#1}}

\newcommand{\forceref}[1]{\raisebox{2pt}{\tikz{\draw[#1, line width=0.9pt](0,0) -- (4.5mm,0);}}}

\newcommand{\PAR}[1]{\vskip4pt \noindent {\bf #1~}}

\def\ie{\emph{i.e}\onedot} 
 
\def\etc{\emph{etc}\onedot} 
 
\def\etal{\emph{et al}\onedot}

\newcolumntype{C}{>{\centering\arraybackslash}X}
\usepackage[breaklinks=true,bookmarks=false]{hyperref}

\cvprfinalcopy

\begin{document}

\title{Revisiting Self-Supervised Visual Representation Learning}

\author{Alexander Kolesnikov\thanks{equal contribution}, Xiaohua Zhai\footnotemark[1], Lucas Beyer\footnotemark[1]\\
Google Brain\\
Z\"urich, Switzerland\\
{\tt\small \{akolesnikov,xzhai,lbeyer\}@google.com}
}

\maketitle

%%%%%%%%% ABSTRACT
\begin{abstract}

Unsupervised visual representation learning remains a largely unsolved problem
in computer vision research.
Among a big body of recently proposed approaches for unsupervised learning of visual representations, a class of self-supervised techniques achieves superior performance on many challenging benchmarks.
A large number of the pretext tasks for self-supervised learning have been studied, but other important aspects, such as the choice of convolutional neural networks (CNN), has not received equal attention.
Therefore, we revisit numerous previously proposed self-supervised models, conduct a thorough large scale study and, as a result, uncover multiple crucial insights.
We challenge a number of common practices in self-supervised visual representation learning and observe that standard recipes for CNN design do not always translate to self-supervised representation learning.
As part of our study, we drastically boost the performance of previously proposed techniques and outperform previously published state-of-the-art results by a large margin.

\end{abstract}

%------------------------------------------------------------------------
\section{Introduction}\label{sec:intro}

Automated computer vision systems have recently made drastic progress.
Many models for tackling challenging tasks such as object recognition,
semantic segmentation or object detection can now compete with humans
on complex visual benchmarks~\cite{he2015delving,xie2017aggregated,he2017mask}.
However, the success of such systems hinges on a large
amount of labeled data, which is not always available and
often prohibitively expensive to acquire.
Moreover, these systems are tailored to specific scenarios, e.g. a model trained on the \emph{ImageNet} (ILSVRC-2012) dataset~\cite{russakovsky2015imagenet} can only recognize 1000 semantic categories or a model that was trained to perceive road traffic at daylight may not work in darkness~\cite{dai2018dark,chen2018domain}.

\begin{figure}[t]
  \begin{center}
    \includegraphics[width=1.0\linewidth]{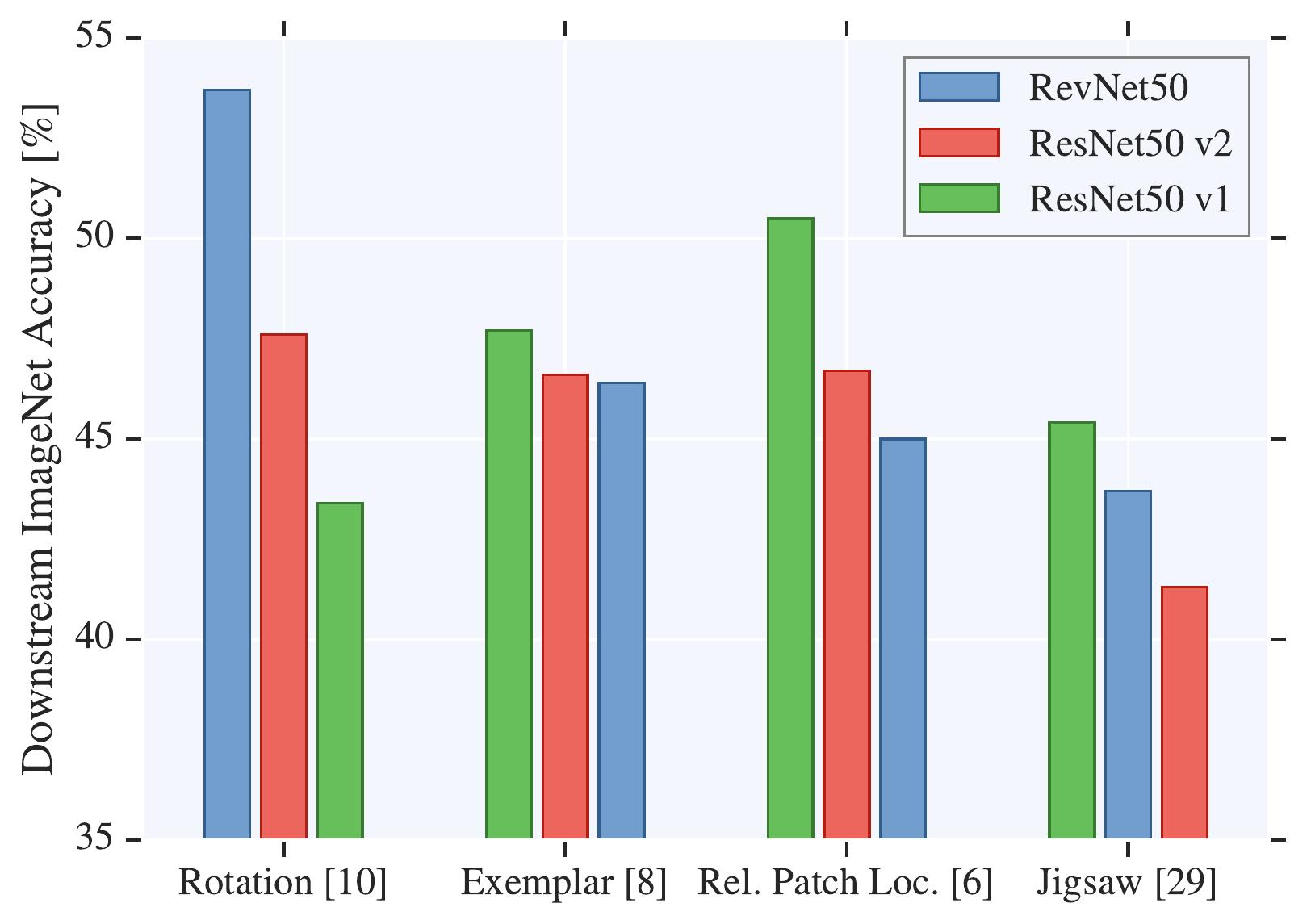}
  \end{center}
  \caption{
     Quality of visual representations learned by various self-supervised learning techniques significantly depends on the convolutional neural network architecture that was used for solving the self-supervised learning task.
     In our paper we provide a large scale in-depth study in support of this observation and discuss its implications for evaluation of self-supervised models.
     }
     \label{fig:teaser}
\end{figure}

As a result, a large research effort is currently focused on systems that
can adapt to new conditions without leveraging a large amount of expensive
supervision.
This effort includes recent advances on transfer learning, domain adaptation,
semi-supervised, weakly-supervised and unsupervised learning.
In this paper, we concentrate on self-supervised visual representation learning, which is a promising sub-class of unsupervised learning.
Self-supervised learning techniques produce state-of-the-art unsupervised representations on standard computer vision benchmarks~\cite{gidaris2018unsupervised,oord2018representation,caron2018deep}.

The self-supervised learning framework requires only unlabeled data in order to formulate a \emph{pretext} learning task such as predicting context~\cite{doersch2015unsupervised} or image rotation~\cite{gidaris2018unsupervised}, for which a target objective can be computed without supervision.
These pretext tasks must be designed in such a way that high-level image understanding is useful for solving them.
As a result, the intermediate layers of convolutional neural networks (CNNs) trained for solving these pretext tasks encode high-level semantic visual representations that are useful for solving \emph{downstream} tasks of interest, such as image recognition.

Most of the prior work, which aims at improving performance of self-supervised
techniques, does so by proposing novel pretext tasks and showing that they
result in improved representations.
Instead, we propose to have a closer look at CNN architectures.
We revisit a prominent subset of the previously proposed pretext tasks and perform a large-scale empirical study using various architectures as base models.
As a result of this study, we uncover numerous crucial insights.
The most important are summarized as follows:
\begin{itemize}
  \item Standard architecture design recipes do not necessarily translate from the fully-supervised to the self-supervised setting. Architecture choices which negligibly affect performance in the fully labeled setting, may significantly affect performance in the self-supervised setting.
  \item In contrast to previous observations with the \emph{AlexNet} architecture~\cite{gidaris2018unsupervised,zhang2017split,noroozi2016unsupervised}, the quality of learned representations in CNN architectures with skip-connections does not degrade towards the end of the model. 
  \item Increasing the number of filters in a CNN model and, consequently, the size of the representation significantly and consistently increases the quality of the learned visual representations.
  \item The evaluation procedure, where a linear model is trained on a fixed visual representation using stochastic gradient descent, is sensitive to the learning rate schedule and may take many epochs to converge.
\end{itemize}
In Section~\ref{sec:core} we present experimental results supporting the above observations and offer additional in-depth insights into the self-supervised learning setting.
We make the code for reproducing our core experimental results publicly available\footnote{\scriptsize\url{https://github.com/google/revisiting-self-supervised}}.

In our study we obtain new state-of-the-art results for visual representations learned without labeled data.
Interestingly, the context prediction~\cite{doersch2015unsupervised} technique that sparked the interest in self-supervised visual representation learning and that serves as the baseline for follow-up research, outperforms all currently published results (among papers on self-supervised learning) if the appropriate CNN architecture is used.

%------------------------------------------------------------------------
\section{Related Work}\label{sec:related_work}

Self-supervision is a learning framework in which a supervised signal for a pretext task is created automatically, in an effort to learn representations that are useful for solving real-world downstream tasks.
Being a generic framework, self-supervision enjoys a wide number of applications, ranging from robotics to image understanding.

In robotics, both the result of interacting with the world, and the fact that multiple perception modalities simultaneously get sensory inputs are strong signals which can be exploited to create self-supervised tasks~\cite{jang2018grasp2vec, sermanet2017time, lee2018making, ebert2018robustness}.

Similarly, when learning representation from videos, one can either make use of the synchronized cross-modality stream of audio, video, and potentially subtitles~\cite{owens2018audio, sayed2018cross, korbar2018cooperative, wiles2018self}, or of the consistency in the temporal dimension~\cite{sermanet2017time}.

In this paper we focus on self-supervised techniques that learn from image databases.
These techniques have demonstrated impressive results for learning high-level image representations.
Inspired by unsupervised methods from the natural language processing domain which rely on predicting words from their context~\cite{mikolov2013efficient}, Doersch \etal \cite{doersch2015unsupervised} proposed a practically successful pretext task of predicting the relative location of image patches.
This work spawned a line of work in patch-based self-supervised visual representation learning methods.
These include a model from~\cite{noroozi2016unsupervised} that predicts the permutation of a ``jigsaw puzzle'' created from the full image and recent follow-ups~\cite{mundhenk2018improvements,noroozi2018boosting}.

In contrast to patch-based methods, some methods generate cleverly designed image-level classification tasks.
For instance, in~\cite{gidaris2018unsupervised} Gidaris \etal propose to randomly rotate an image by one of four possible angles and let the model predict that rotation.
Another way to create class labels is to use clustering of the images~\cite{caron2018deep}. 
Yet another class of pretext tasks contains tasks with dense spatial outputs.
Some prominent examples are image inpainting~\cite{pathakCVPR16context}, image colorization~\cite{zhang2016colorful}, its improved variant split-brain~\cite{zhang2017split} and motion segmentation prediction~\cite{pathak2017learning}.
Other methods instead enforce structural constraints on the representation space.
Noroozi~\etal propose an equivariance relation to match the sum of multiple tiled representations to a single scaled representation~\cite{noroozi2017representation}.
Authors of~\cite{oord2018representation} propose to predict future patches in representation space via autoregressive predictive coding.

Our work is complimentary to the previously discussed methods, which introduce new pretext tasks, since we show how existing self-supervision methods can significantly benefit from our insights.

Finally, many works have tried to combine multiple pretext tasks in one way or another.
For instance, Kim \etal extend the ``jigsaw puzzle'' task by combining it with colorization and inpainting in~\cite{kim2018learning}.
Combining the jigsaw puzzle task with clustering-based pseudo labels as in~\cite{caron2018deep} leads to the method called Jigsaw++~\cite{noroozi2018boosting}.
Doersch and Zisserman~\cite{doersch2017multi} implement four different self-supervision methods and make one single neural network learn all of them in a multi-task setting.

The latter work is similar to ours since it contains a comparison of different self-supervision methods using a unified neural network architecture, but with the goal of combining all these tasks into a single self-supervision task.
The authors use a modified ResNet101 architecture~\cite{he2016deep} without further investigation and explore the combination of multiple tasks, whereas our focus lies on investigating the influence of architecture design on the representation quality.

%------------------------------------------------------------------------
\section{Self-supervised study setup}\label{sec:eval}

In this section we describe the setup of our study and motivate our key choices.
We begin by introducing six CNN models in Section~\ref{subsec:cnn_models} and proceed by describing the four self-supervised learning approaches used in our study in Section~\ref{subsec:methods}.
Subsequently, we define our evaluation metrics and datasets in Sections~\ref{subsec:eval}~and~\ref{subsec:datasets}. 
Further implementation details can be found in Supplementary Material.

%------------------------------------------------------------------------
\subsection{Architectures of CNN models}\label{subsec:cnn_models}

A large part of the self-supervised techniques for visual representation approaches uses \emph{AlexNet}~\cite{KrizhevskyNIPS12} architecture.
In our study, we investigate whether the landscape of self-supervision techniques changes when using modern network architectures.
Thus, we employ variants of \emph{ResNet} and a batch-normalized \emph{VGG} architecture, all of which achieve high performance in the fully-supervised training setup.
\emph{VGG} is structurally close to \emph{AlexNet} as it does not have skip-connections and uses fully-connected layers.

In our preliminary experiments, we observed an intriguing property of ResNet models:
the quality of the representations they learn does not degrade towards the end of the network (see Section~\ref{subsec:layers}).
We hypothesize that this is a result of skip-connections making residual units invertible under certain circumstances~\cite{behrmann2018invertible}, hence facilitating the preservation of information across the depth even when it is irrelevant for the pretext task.
Based on this hypothesis, we include \emph{RevNet}s~\cite{GomezNIPS17} into our study, which come with stronger invertibility guarantees while being structurally similar to ResNets.

\PAR{ResNet} was introduced by He~\etal~\cite{he2016deep}, and we use the width-parametrization proposed in~\cite{zagoruyko2016wide}:
the first $7\times7$ convolutional layer outputs $16 \times k$ channels, where $k$ is the \emph{widening factor}, defaulting to 4.
This is followed by a series of \emph{residual units} of the form $y := x + \mathcal{F}(x)$, where $\mathcal{F}$ is a \emph{residual function} consisting of multiple convolutions, ReLU non-linearities~\cite{nair2010rectified} and batch normalization layers~\cite{ioffe2015batch}.
The variant we use, \emph{ResNet50}, consists of four \emph{blocks} with 3, 4, 6, and 3 such units respectively, and we refer to the output of each block as \emph{block1}, \emph{block2}, \etc.
The network ends with a global spatial average pooling producing a vector of size $512 \times k$, which we call \emph{pre-logits} as it is followed only by the final, task-specific \emph{logits} layer.
More details on this architecture are provided in~\cite{he2016deep}.

In our experiments we explore $k \in \{4, 8, 12, 16\}$, resulting in \emph{pre-logits} of size $2048$, $4096$, $6144$ and $8192$ respectively.
For some self-supervised techniques we skip configurations that do not fit into memory.

Moreover, we analyze the sensitivity of the self-supervised setting to underlying architectural details by using two variants of ordering operations known as \emph{ResNet~v1}~\cite{he2016deep} and \emph{ResNet~v2}~\cite{he2016identity} as well as a variant without ReLU preceding the global average pooling, which we mark by a ``(-)''.
Notably, these variants perform similarly on the pretext task.

\PAR{RevNet} slightly modifies the design of the residual unit such that it becomes analytically invertible~\cite{GomezNIPS17}.
We note that the residual unit used in~\cite{GomezNIPS17} is equivalent to double application of the residual unit from~\cite{JacobsenICLR18} or~\cite{DinhICLR17}.
Thus, for conceptual simplicity, we employ the latter type of unit, which can be defined as follows.
The input $x$ is split channel-wise into two equal parts $x_1$ and $x_2$.
The output $y$ is then the concatenation of $y_2 := x_2$ and $y_1 := x_1 + \mathcal{F}(x_2)$.

It easy to see that this residual unit is invertible, because its inverse can be computed in closed form as
$x_2 = y_2$ and $x_1 = y_1 - \mathcal{F}(x_2)$.

Apart from this slightly different residual unit, \emph{RevNet} is structurally identical to \emph{ResNet} and thus we use the same overall architecture and nomenclature for both.
In our experiments we use \emph{RevNet50} network, that has the same depth and number of channels as the
original \emph{Resnet50} model.
In the fully labelled setting, \emph{RevNet} performs only marginally worse than its architecturally equivalent \emph{ResNet}.

\PAR{VGG} as proposed in~\cite{simonyan2014very} consists of a series of $3\times3$ convolutions followed by ReLU non-linearities, arranged into \emph{blocks} separated by max-pooling operations.
The \emph{VGG19} variant we use has 5 such blocks of 2, 2, 4, 4, and 4 convolutions respectively.
We follow the common practice of adding batch normalization between the convolutions and non-linearities.

In an effort to unify the nomenclature with ResNets, we introduce the \emph{widening factor} $k$ such that $k=8$ corresponds to the architecture in~\cite{simonyan2014very}, \ie the initial convolution produces $8\times k$ channels and the fully-connected layers have $512\times k$ channels.
Furthermore, we call the inputs to the second, third, fourth, and fifth max-pooling operations \emph{block1} to \emph{block4}, respectively, and the input to the last fully-connected layer \emph{pre-logits}.

%------------------------------------------------------------------------
\subsection{Self-supervised techniques}\label{subsec:methods}

In this section we describe the self-supervised techniques that are used in our study.

\PAR{Rotation}\cite{gidaris2018unsupervised}: Gidaris~\etal propose to produce 4 copies of a single image by rotating it by \{0\degree, 90\degree, 180\degree, 270\degree\} and let a single network predict the rotation which was applied---a 4-class classification task.
Intuitively, a good model should learn to recognize canonical orientations of objects in natural images.

\PAR{Exemplar}\cite{dosovitskiy2014exemplar}:
In this technique, every individual image corresponds to its own class, and multiple examples of it are generated by heavy random data augmentation such as translation, scaling, rotation, and contrast and color shifts. We use data augmentation mechanism from~\cite{szegedy2015going}.
\cite{doersch2017multi} proposes to use the triplet loss~\cite{schroff2015facenet,HermansBeyer2017Arxiv} in order to scale this pretext task to a large number of images (hence, classes) present in the ImageNet dataset.
The triplet loss avoids explicit class labels and, instead, encourages examples of the same image to have representations that are close in the Euclidean space while also being far from the representations of different images.
Example representations are given by a 1000-dimensional \emph{logits} layer.

\PAR{Jigsaw}\cite{noroozi2016unsupervised}:
the task is to recover relative spatial position of 9 randomly sampled image patches after a random permutation of these patches was performed.
All of these patches are sent through the same network, then their representations from the \emph{pre-logits} layer are concatenated and passed through a two hidden layer fully-connected multi-layer perceptron (MLP), which needs to predict a permutation that was used.
In practice, the fixed set of 100 permutations from~\cite{noroozi2016unsupervised} is used.

In order to avoid shortcuts relying on low-level image statistics such as chromatic aberration~\cite{noroozi2016unsupervised} or edge alignment, patches are sampled with a random gap between them. Each patch is then independently 
converted to grayscale with probability \sfrac{2}{3} and
normalized to zero mean and unit standard deviation.
More details on the preprocessing are provided in Supplementary Material.
After training, we extract representations by averaging the representations of nine uniformly sampled, colorful, and normalized patches of an image.

\PAR{Relative Patch Location}\cite{doersch2015unsupervised}:
The pretext task consists of predicting the relative location of two given patches of an image.
The model is similar to the Jigsaw one, but in this case the 8 possible relative spatial relations between two patches need to be predicted, e.g. ``below'' or ``on the right and above''.
We use the same patch prepossessing as in the Jigsaw model and also extract
final image representations by averaging representations of 9 cropped patches.

%------------------------------------------------------------------------
\subsection{Evaluation of Learned Visual Representations}\label{subsec:eval}

We follow common practice and evaluate the learned visual representations
by using them for training a linear logistic regression model to solve multiclass image classification tasks requiring high-level scene understanding.
These tasks are called \emph{downstream tasks}.
We extract the representation from the (frozen) network at the \emph{pre-logits} level, but investigate other possibilities in Section~\ref{subsec:layers}.

In order to enable fast evaluation, we use an efficient convex optimization technique for training the logistic regression model unless specified otherwise.
Specifically, we precompute the visual representation for all training images and train the logistic regression using L-BFGS~\cite{liu1989limited}.

For consistency and fair evaluation, when comparing to the prior literature in Table~\ref{tbl:bigtable_imagenet}, we opt for using stochastic gradient descent (SGD) with momentum and use data augmentation during training.

We further investigate this common evaluation scheme in Section~\ref{subsec:nonlinear}, where we use a more expressive model, which is an MLP with a single hidden layer with 1000 channels and the ReLU non-linearity after it.
More details are given in Supplementary material.
%

%------------------------------------------------------------------------
\subsection{Datasets}\label{subsec:datasets}

In our experiments, we consider two widely used  image classification datasets: \emph{ImageNet} and \emph{Places205}.

\emph{ImageNet} contains roughly \SI{1.3}{\text{million}} natural images that represent 1000 various semantic classes.
There are $50\,000$ images in the official validation and test sets, but since the official test set is held private, results in the literature are reported on the validation set.
In order to avoid overfitting to the official validation split, we report numbers on our own validation split ($50\,000$ random images from the training split) for all our studies except in Table~\ref{tbl:sota}, where for a fair comparison with the literature we evaluate on the official validation set.

The \emph{Places205} dataset consists of roughly \SI{2.5}{\text{million}} images depicting 205 different scene types such as \emph{airfield}, \emph{kitchen}, \emph{coast}, etc.
This dataset is qualitatively different from \emph{ImageNet} and, thus, a good candidate for evaluating how well the learned representations generalize to new unseen data of different nature.
We follow the same procedure as for \emph{ImageNet} regarding validation splits for the same reasons.

%------------------------------------------------------------------------
%------------------------------------------------------------------------
\section{Experiments and Results}\label{sec:core}

In this section we present and interpret results of our large-scale study.
All self-supervised models are trained on \emph{ImageNet}  (without labels) and consequently evaluated on our own hold-out validation splits of \emph{ImageNet} and \emph{Places205}.
Only in Table~\ref{tbl:sota}, when we compare to the results from the prior literature, we use the official \emph{ImageNet} and \emph{Places205} validation splits.

\begin{table*}[t]
  \caption{%
  Evaluation of representations from self-supervised techniques based on various CNN architectures.
  The scores are accuracies (in \%) of a linear logistic regression model trained on top of these representations 
  using \emph{ImageNet} training split. Our validation split is used for computing accuracies.
  The architectures marked by a ``(-)'' are slight variations described in Section~\ref{subsec:cnn_models}.
  Sub-columns such as $4\times$ correspond to widening factors.
  Top-performing architectures in a column are bold; the best pretext task for each model is underlined.}
  \label{tbl:bigtable_imagenet}
  \setlength{\tabcolsep}{0pt}
  \setlength{\extrarowheight}{5pt}
  \renewcommand{\arraystretch}{0.75}
  \centering
  \begin{tabularx}{\linewidth}{p{2.7cm}CCCCp{7pt}CCCp{7pt}CCp{7pt}CC}
    \toprule[1pt]
    \multirow{3}{*}{Model} & \multicolumn{4}{c}{Rotation} && \multicolumn{3}{c}{Exemplar} && \multicolumn{2}{c}{RelPatchLoc} && \multicolumn{2}{c}{Jigsaw}\\
    \cmidrule[0.5pt]{2-5} \cmidrule[0.5pt]{7-9} \cmidrule[0.5pt]{11-12} \cmidrule[0.5pt]{14-15}
     & $4\times$ & $8\times$ & $12\times$ & $16\times$ && $4\times$ & $8\times$ & $12\times$ && $4\times$ & $8\times$ && $4\times$ & $8\times$\\

    \midrule

    RevNet50 & \bestc{47.3} & \bestc{50.4} & \bestc{53.1} & \bestr{\bestc{53.7}} && \bestc{42.4} & 45.6 & 46.4 && 40.6 & 45.0 && 40.1 & 43.7 \\
    ResNet50 v2 & 43.8 & 47.5 & 47.2 & \bestr{47.6} && \bestc{43.0} & 45.7 & 46.6 && 42.2 & 46.7 && 38.4 & 41.3 \\
    ResNet50 v1 & 41.7 & 43.4 & 43.3 & 43.2 && \bestc{42.8} & \bestc{46.9} & \bestc{47.7} && \bestc{46.8} & \bestr{\bestc{50.5}} && \bestc{42.2} & \bestc{45.4} \\

    \arrayrulecolor{lightgray}\midrule[0.25pt]\arrayrulecolor{black}
    RevNet50 (-) & 45.2 & \bestc{51.0} & \bestc{52.8} & \bestr{\bestc{53.7}} && 38.0 & 42.6 & 44.3 && 33.8 & 43.5 && 36.1 & 41.5 \\
    ResNet50 v2 (-) & 38.6 & 44.5 & 47.3 & \bestr{48.2} && 33.7 & 36.7 & 38.2 &&  38.6 & 43.4 && 32.5 & 34.4 \\
    \arrayrulecolor{lightgray}\midrule[0.25pt]\arrayrulecolor{black}
    VGG19-BN & 16.8 & 14.6 & 16.6 & 22.7 && 26.4 & 28.3 & \bestr{29.0} && 28.5 & \bestr{29.4} && 19.8 & 21.1 \\
    \bottomrule
  \end{tabularx}
\end{table*}

%------------------------------------------------------------------------
\subsection{Evaluation on \emph{ImageNet} and \emph{Places205}}\label{subsec:bigtable}

In Table~\ref{tbl:bigtable_imagenet} we highlight our main evaluation results:
we measure the representation quality produced by six different CNN architectures with various widening factors (Section~\ref{subsec:cnn_models}), trained using four self-supervised learning techniques (Section~\ref{subsec:methods}).
We use the \emph{pre-logits} of the trained self-supervised networks as representation.
We follow the standard evaluation protocol (Section~\ref{subsec:eval}) which measures representation quality as the accuracy of a linear regression model trained and evaluated on the \emph{ImageNet} dataset.

\begin{figure}[b]
  \begin{center}
    \includegraphics[width=1.0\linewidth]{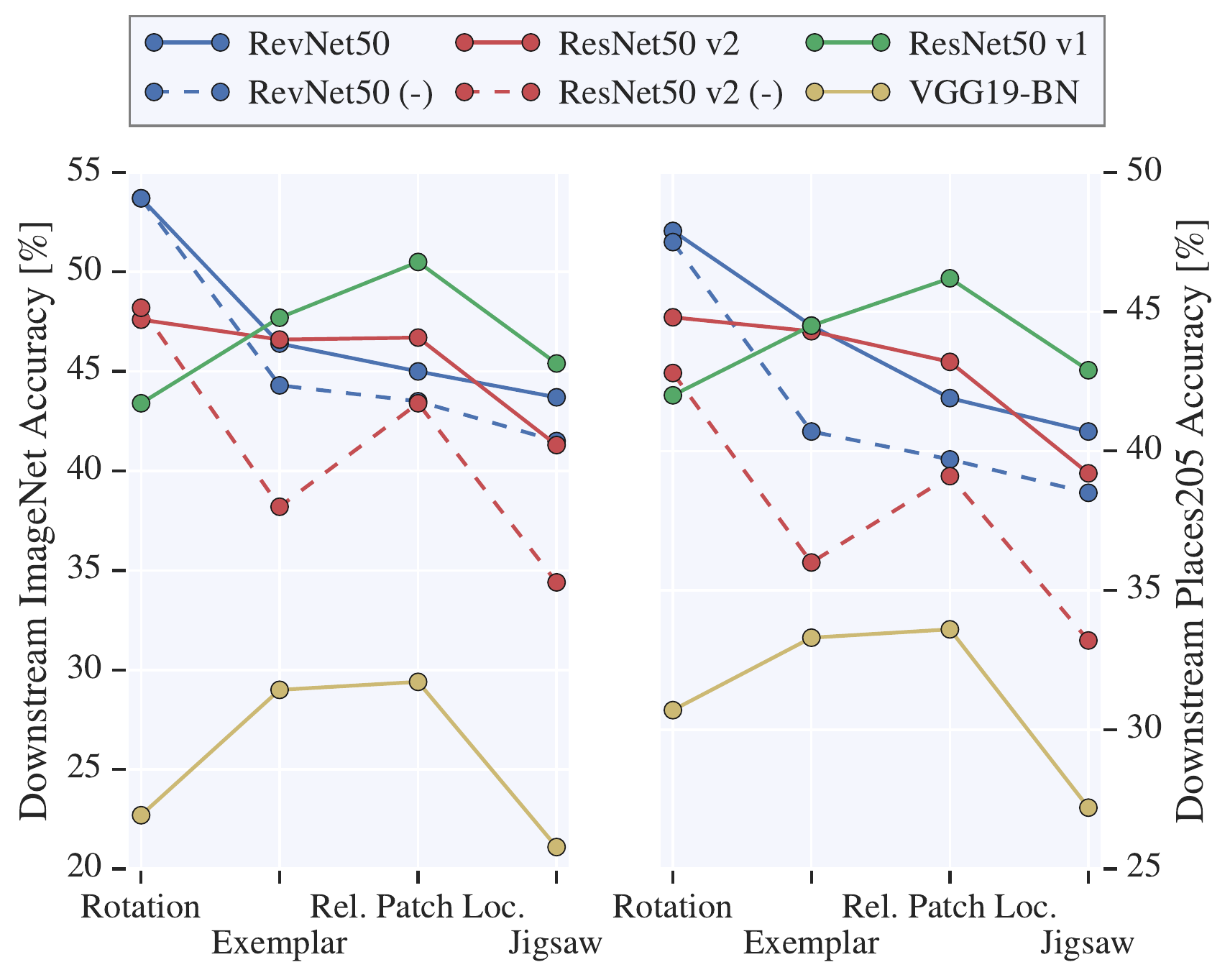}
  \end{center}
  \caption{
     Different network architectures perform significantly differently across self-supervision tasks. This observation generalizes across datasets: \emph{ImageNet} evaluation is shown on the left and \emph{Places205} is shown on the right.
  }\label{fig:imagenet_vs_places}
\end{figure}

\begin{table}[b]
  \setlength{\tabcolsep}{0pt}
  \setlength{\extrarowheight}{5pt}
  \renewcommand{\arraystretch}{0.75}
  \centering
  \begin{tabularx}{\linewidth}{p{0.4cm}p{1pt}p{3.7cm}CCp{7pt}CC}
    \toprule[1pt]
     \multirow{3}{*}{\rotatebox{90}{\hspace*{5pt}Family}} && & \multicolumn{2}{c}{ImageNet} && \multicolumn{2}{c}{Places205}\\
    \cmidrule[0.5pt]{4-5} \cmidrule[0.5pt]{7-8}
     && & Prev. & Ours && Prev. & Ours\\
    \midrule

    A && Rotation\cite{gidaris2018unsupervised} & 38.7 & \textbf{55.4} && 35.1 & \textbf{48.0} \\
    R && Exemplar\cite{doersch2017multi} & 31.5 & 46.0 && - & 42.7 \\
    R && Rel. Patch Loc.\cite{doersch2017multi} & 36.2 & 51.4 && - & 45.3 \\
    A && Jigsaw\cite{noroozi2016unsupervised,zhang2017split} & 34.7 & 44.6 && 35.5 & 42.2 \\

    \arrayrulecolor{lightgray}\midrule[0.25pt]\arrayrulecolor{black}

    V && CC+vgg-Jigsaw++\cite{noroozi2018boosting} & 37.3 & - && 37.5 & - \\
    A && Counting\cite{noroozi2017representation}  & 34.3 & - && 36.3 & - \\
    A && Split-Brain\cite{zhang2017split}          & 35.4 & - && 34.1 & - \\
    V && DeepClustering\cite{caron2018deep}        & \textbf{41.0} & - && \textbf{39.8} & - \\

    \arrayrulecolor{lightgray}\midrule[0.25pt]\arrayrulecolor{black}

    R && CPC\cite{oord2018representation} & 48.7\makebox[0pt]{\hspace{5pt}$^\dagger$} & - && - & - \\

    \arrayrulecolor{lightgray}\midrule[0.25pt]\arrayrulecolor{black}

    R && Supervised RevNet50    & 74.8 & 74.4 &&   -  & 58.9 \\
    R && Supervised ResNet50 v2 & 76.0 & 75.8 &&   -  & 61.6 \\
    V && Supervised VGG19       & 72.7 & 75.0 && 58.9 & 61.5 \\
    \bottomrule
    \multicolumn{8}{r}{\emph{\footnotesize{$^\dagger$ marks results reported in unpublished manuscripts.}}}
  \end{tabularx}
  \vspace*{3pt}
  \caption{
    Comparison of the published self-supervised models to our best models.
    The scores correspond to accuracy of linear logistic regression that is trained on top of representations provided by self-supervised models.
    Official validation splits of \emph{ImageNet} and \emph{Places205} are used for computing accuracies.
    The ``Family'' column shows which basic model architecture was used in the referenced literature: \textbf{A}lexNet, \textbf{V}GG-style, or \textbf{R}esidual.}  \label{tbl:sota}
\end{table}

Now we discuss key insights that can be learned from the table and motivate our further in-depth analysis.
First, we observe that similar models often result in visual representations that have significantly different performance.
Importantly, \textbf{neither is the ranking of architectures consistent across different methods, nor is the ranking of methods consistent across architectures}.
For instance, the \emph{RevNet50 v2} model excels under \emph{Rotation} self-supervision, but is not the best model in other scenarios.
Similarly, \emph{relative patch location} seems to be the best method when basing the comparison on the \emph{ResNet50 v1} architecture, but not otherwise.
Notably, \emph{VGG19-BN} consistently demonstrates the worst performance,
even though it achieves performance similar to \emph{ResNet50} models on standard vision benchmarks~\cite{simonyan2014very}.
Note that VGG19-BN performs better when using representations from layers earlier than the \emph{pre-logit} layer are used, though still falls short. We investigate this in Section~\ref{subsec:layers}.
We depict the performance of the models with the largest widening factor in Figure~\ref{fig:imagenet_vs_places} (left), which displays these ranking inconsistencies.

Our second observation is that increasing the number of channels in CNN models improves performance of self-supervised models.
While this finding is in line with the fully-supervised setting~\cite{zagoruyko2016wide}, we note that the benefit is more pronounced in the context of self-supervised representation learning, a fact not yet acknowledged in the literature.

We further evaluate how visual representations trained in a self-supervised manner on \emph{ImageNet} generalize to other datasets.
Specifically, we evaluate all our models on the Places205 dataset using the same evaluation protocol.
The performance of models with the largest widening factor are reported in Figure~\ref{fig:imagenet_vs_places} (right) and the full result table is provided in Supplementary Material.
We observe the following pattern: ranking of models evaluated on \emph{Places205} is consistent with that of models evaluated on \emph{ImageNet}, indicating that our findings generalize to new datasets.

%------------------------------------------------------------------------
\subsection{Comparison to prior work}

In order to put our findings in context, we select the best model for each self-supervision from Table~\ref{tbl:bigtable_imagenet}
and compare them to the numbers reported in the literature.
For this experiment only, we precisely follow standard protocol by training the linear model with stochastic gradient descent (SGD) on the full \emph{ImageNet} training split and evaluating it on the public validation set of both \emph{ImageNet} and \emph{Places205}.
We note that in this case the learning rate schedule of the evaluation plays an important role, which we elaborate in Section~\ref{sec:sgd-long}.

\begin{figure}[t]
  \begin{center}
    \includegraphics[width=1.0\linewidth]{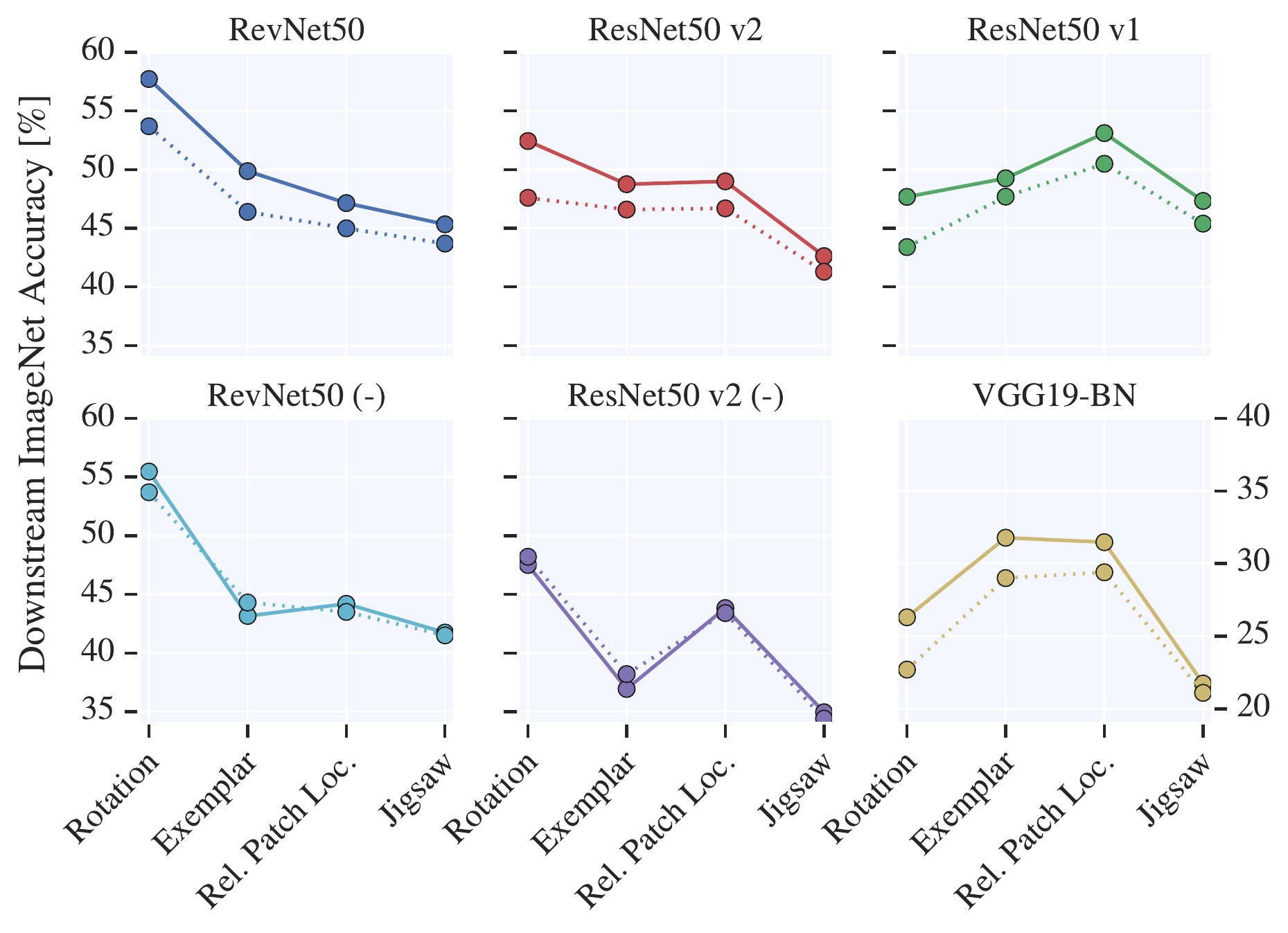}
  \end{center}
  \caption{
     Comparing linear evaluation (\protect\forceref{black,dotted}) of the representations to non-linear (\protect\forceref{black}) evaluation, \ie training a multi-layer perceptron instead of a linear model.
     Linear evaluation is not limiting: conclusions drawn from it carry over to the non-linear evaluation.
  }\label{fig:linear_vs_mlp}
\end{figure}

Table~\ref{tbl:sota} summarizes our results.
Surprisingly, as a result of selecting the right architecture for each self-supervision and increasing the widening factor, our models significantly outperform previously reported results.
Notably, context prediction~\cite{doersch2015unsupervised}, one of the earliest published methods, achieves \SI{51.4}{\percent} top-1 accuracy on \emph{ImageNet}.
Our strongest model, using \emph{Rotation}, attains unprecedentedly high accuracy
of \SI{55.4}{\percent}.
Similar observations hold when evaluating on \emph{Places205}.

Importantly, our design choices result in almost halving the gap between previously published self-supervised result and fully-supervised results on two standard benchmarks.
Overall, these results reinforce our main insight that in self-supervised learning architecture choice matters as much as choice of a pretext task.
%------------------------------------------------------------------------
\subsection{A linear model is adequate for evaluation.}
\label{subsec:nonlinear}

Using a linear model for evaluating the quality of a representation requires that the information relevant to the evaluation task is linearly separable in representation space.
This is not necessarily a prerequisite for a ``useful'' representation.
Furthermore, using a more powerful model in the evaluation procedure might make the architecture choice for a self-supervised task less important.
Hence, we consider an alternative evaluation scenario where we use a multi-layer perceptron (MLP) for solving the evaluation task, details of which are provided in Supplementary Material.

Figure~\ref{fig:linear_vs_mlp} clearly shows that the MLP provides only marginal improvement over the linear evaluation and the relative performance of various settings is mostly unchanged.
We thus conclude that the linear model is adequate for evaluation purposes.

%------------------------------------------------------------------------
\subsection{Better performance on the pretext task does not always translate to better representations.}
\label{subsec:updown}

\begin{figure}[t]
  \begin{center}
    \includegraphics[width=1.0\linewidth]{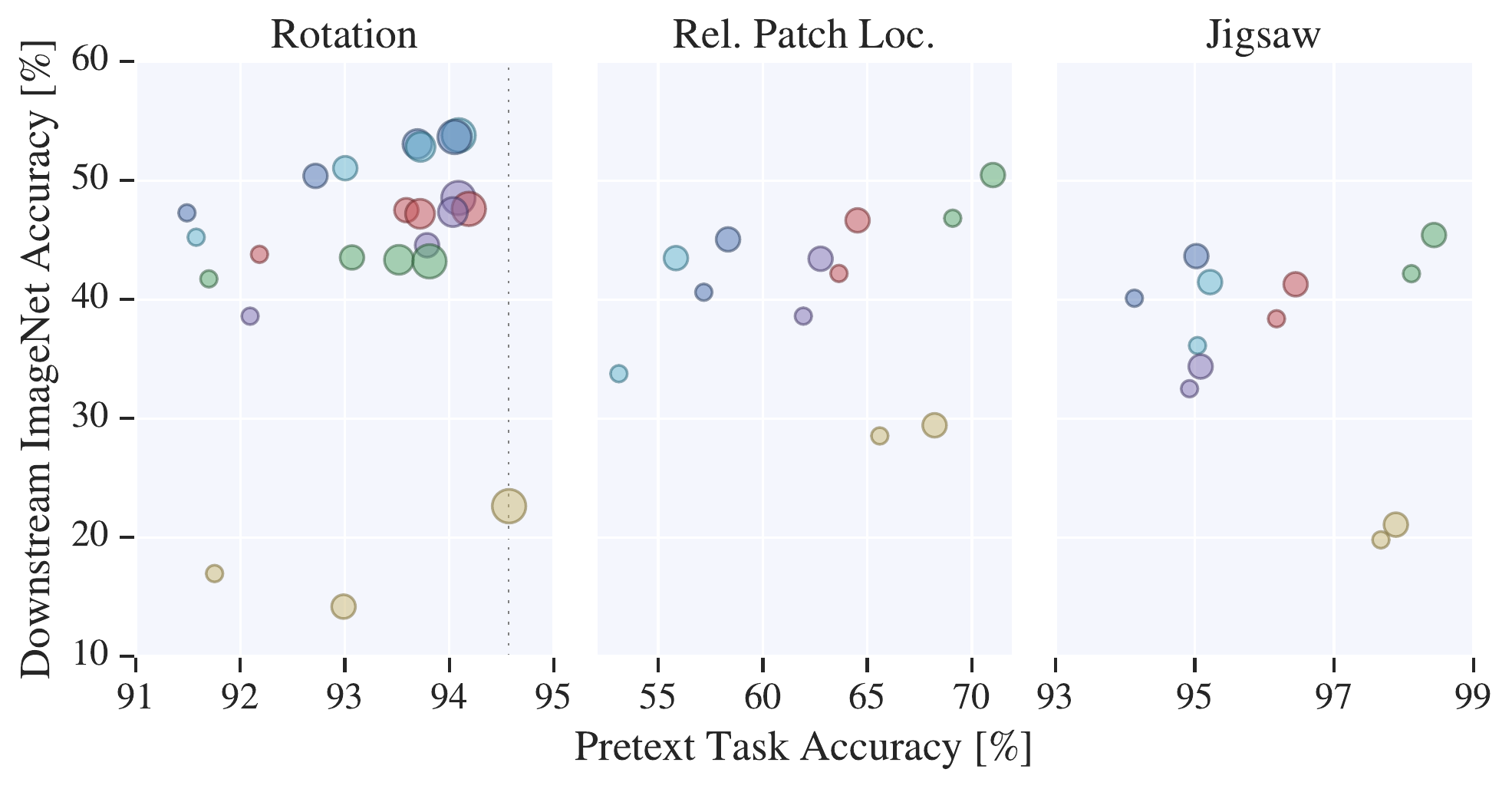}
  \end{center}
  \caption{
     A look at how predictive pretext performance is to eventual downstream performance.
     Colors correspond to the architectures in Figure~\ref{fig:linear_vs_mlp} and circle size to the widening factor $k$.
     Within an architecture, pretext performance is somewhat predictive, but it is not so across architectures.
     For instance, according to pretext accuracy, the widest \emph{VGG} model is the best one for \emph{Rotation}, but it performs poorly on the downstream task.
  }\label{fig:updown}
\end{figure}

In many potential applications of self-supervised methods, we do not have access to downstream labels for evaluation.
In that case, how can a practitioner decide which model to use?
Is performance on the pretext task a good proxy?

\begin{figure*}[t]
  \begin{center}
    \includegraphics[width=1.0\linewidth]{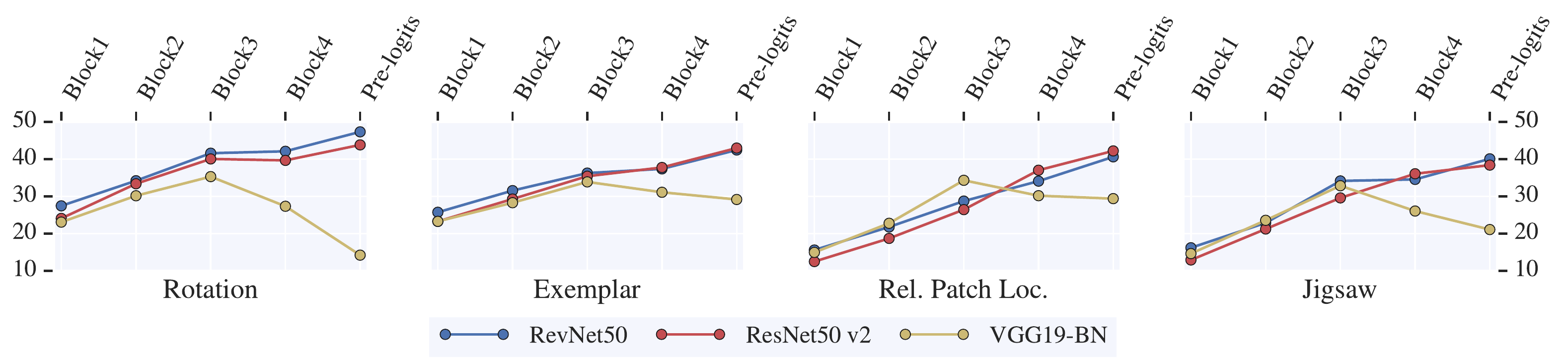}
  \end{center}
  \caption{
     Evaluating the representation from various depths within the network.
     The vertical axis corresponds to downstream ImageNet performance in percent.
     For residual architectures, the \emph{pre-logits} are always best.
  }\label{fig:blocks}
\end{figure*}

In Figure~\ref{fig:updown} we plot the performance on the pretext task against the evaluation on \emph{ImageNet}.
It turns out that performance on the pretext task is a good proxy only once the model architecture is fixed, but it can unfortunately not be used to reliably select the model architecture.
Other label-free mechanisms for model-selection need to be devised, which we believe is an important and underexplored area for future work.

%------------------------------------------------------------------------
\subsection{Skip-connections prevent degradation of representation quality towards the end of CNNs.}
\label{subsec:layers}

We are interested in how representation quality depends on the layer choice and how skip-connections affect this dependency.
Thus, we evaluate representations from five intermediate layers in three models: \emph{Resnet v2}, \emph{RevNet} and \emph{VGG19-BN}.
The results are summarized in Figure~\ref{fig:blocks}.

\begin{figure}[t]
  \begin{center}
    \includegraphics[width=1.0\linewidth]{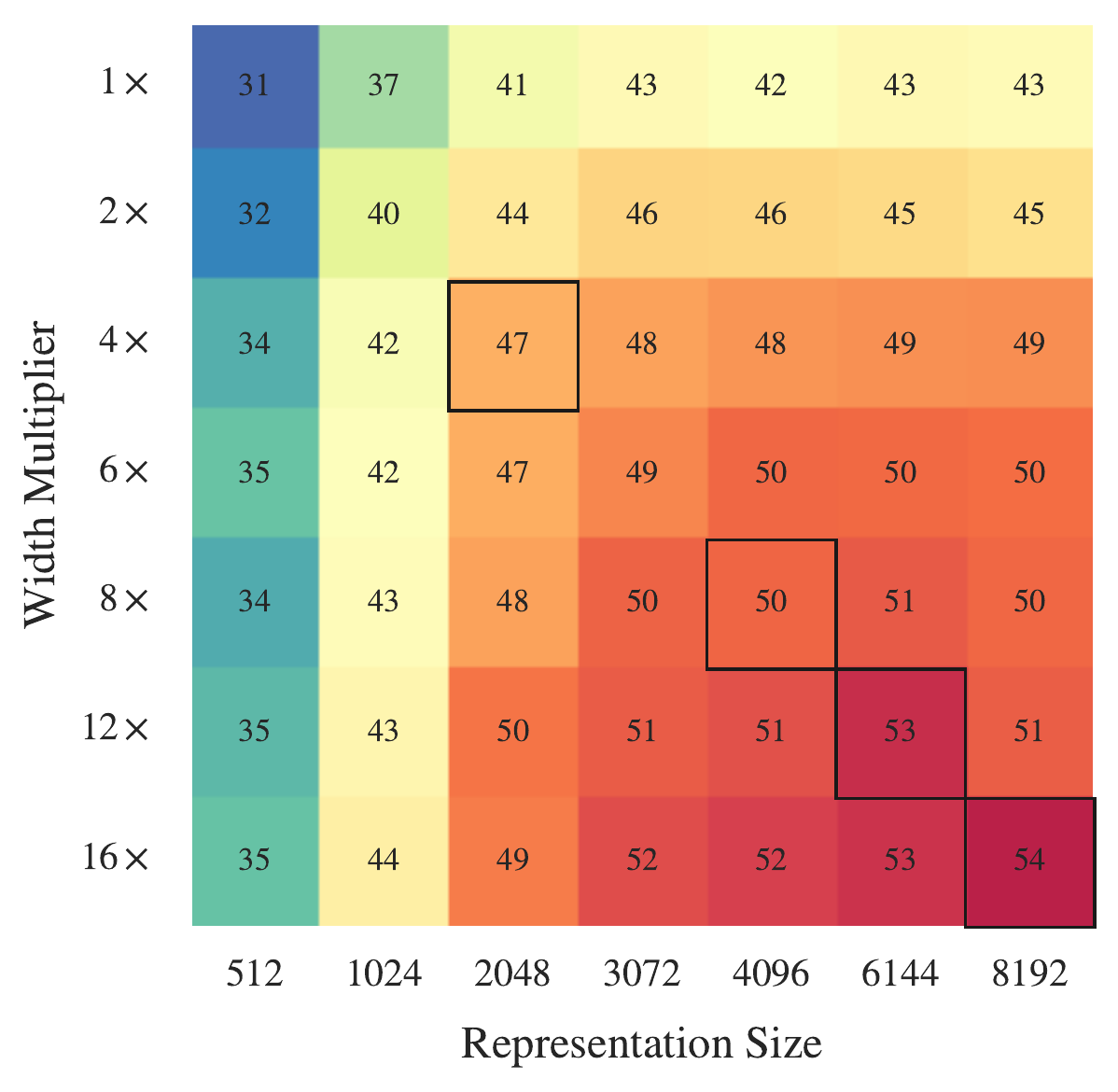}
  \end{center}
  \caption{
     Disentangling the performance contribution of network widening factor versus representation size. Both matter independently, and larger is always better. Scores are accuracies of logistic regression on \emph{ImageNet}. Black squares mark models which are also present in Table~\ref{tbl:bigtable_imagenet}.
  }\label{fig:width_vs_repsize}
\end{figure}

Similar to prior observations~\cite{gidaris2018unsupervised,zhang2017split,noroozi2016unsupervised} for \emph{AlexNet}~\cite{krizhevsky2012imagenet}, the quality of representations in \emph{VGG19-BN} deteriorates towards the end of the network.
We believe that this happens because the models specialize to the pretext task in the later layers and, consequently, discard more general semantic features
present in the middle layers.

In contrast, we observe that this is not the case for models with skip-connections:
representation quality in \emph{ResNet} consistently increases up to the final \emph{pre-logits} layer.
We hypothesize that this is a result of \emph{ResNet}'s residual units being
invertible under some conditions~\cite{behrmann2018invertible}.
Invertible units preserve all information learned in intermediate layers, and, thus, prevent deterioration of representation quality.

We further test this hypothesis by using the \emph{RevNet} model that has stronger invertibility guarantees.
Indeed, it boosts performance by more than \SI{5}{\percent} on the \emph{Rotation} task, albeit it does not result in improvements across other tasks.
We leave identifying further scenarios where \emph{Revnet} models result in significant boost of performance for the future research.

%------------------------------------------------------------------------
\subsection{Model width and representation size strongly influence the representation quality.}
\label{subsec:width_vs_size}

Table~\ref{tbl:bigtable_imagenet} shows that using a wider network architecture consistently leads to better representation quality.
It should be noted that increasing the network's width has the side-effect of also increasing the dimensionality of the final representation (Section~\ref{subsec:cnn_models}).
Hence, it is unclear whether the increase in performance is due to increased network capacity or to the use of higher-dimensional representations, or to the interplay of both.

In order to answer this question, we take the best rotation model (RevNet50) and disentangle the network width from the representation size by adding an additional linear layer to control the size of the \emph{pre-logits} layer.
We then vary the widening factor and the representation size independently of each other, training each model from scratch on \emph{ImageNet} with the \emph{Rotation} pretext task.
The results, evaluated on the \emph{ImageNet} classification task, are shown in Figure~\ref{fig:width_vs_repsize}.
In essence, it is possible to increase performance by increasing either model capacity, or representation size, but increasing both jointly helps most.
Notably, one can significantly boost performance of a very thin model from \SI{31}{\percent} to \SI{43}{\percent} by increasing representation size.

%------------------------------------------------------------------------
\begin{figure}[t]
  \begin{center}
    \includegraphics[width=1.0\linewidth]{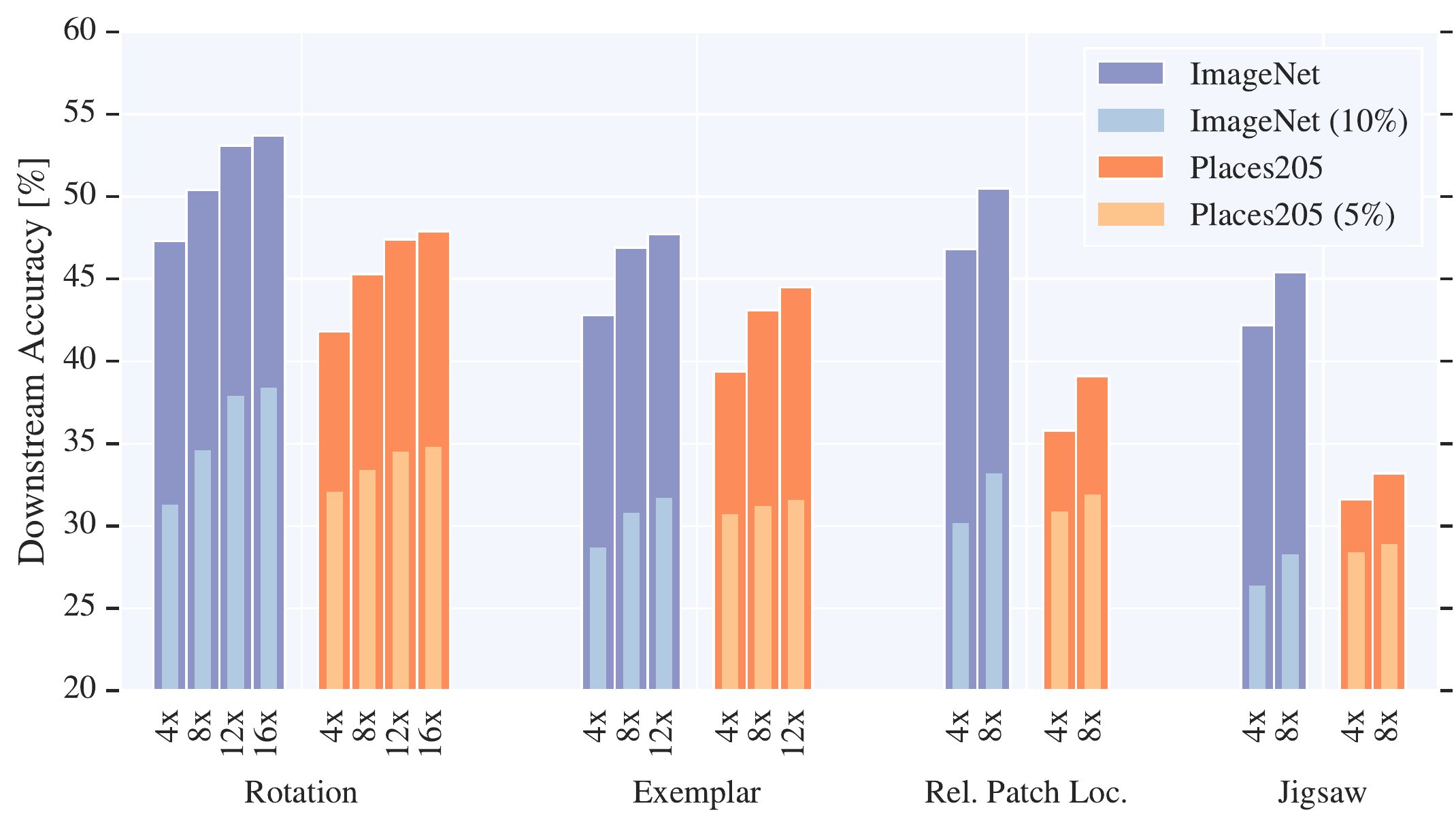}
  \end{center}
  \caption{
     Performance of the best models evaluated using all data as well as a subset of the data.
     The trend is clear: increased widening factor increases performance across the board.
  }\label{fig:ff_small_data}
\end{figure}

\PAR{Low-data regime.}
In principle, the effectiveness of increasing model capacity and representation size might only work on relatively large datasets for downstream evaluation, and might hurt representation usefulness in the low-data regime.
In Figure~\ref{fig:ff_small_data}, we depict how the number of channels affects the evaluation using both full and heavily subsampled (\SI{10}{\percent} and \SI{5}{\percent}) \emph{ImageNet} and \emph{Places205} datasets.

We observe that increasing the widening factor consistently boosts performance in both the full- and low-data regimes.
We present more low-data evaluation experiments in Supplementary Material.
This suggests that self-supervised learning techniques are likely to benefit from using CNNs with increased number of channels across wide range of scenarios.

\subsection{SGD for training linear model takes long time to converge}\label{sec:sgd-long}

In this section we investigate the importance of the SGD optimization schedule for training logistic regression in downstream tasks.
We illustrate our findings for linear evaluation of the \emph{Rotation} task, others behave the same and are provided in Supplementary Material.

We train the linear evaluation models with a mini-batch size of 2048 and an initial learning rate of 0.1, which we decay twice by a factor of 10.
Our initial experiments suggest that when the first decay is made has a large influence on the final accuracy.
Thus, we vary the moment of first decay, applying it after 30, 120 or 480 epochs.
After this first decay, we train for an extra 40 extra epochs, with a second decay after the first 20.

\begin{figure}[t]
  \begin{center}
    \includegraphics[width=1.0\linewidth]{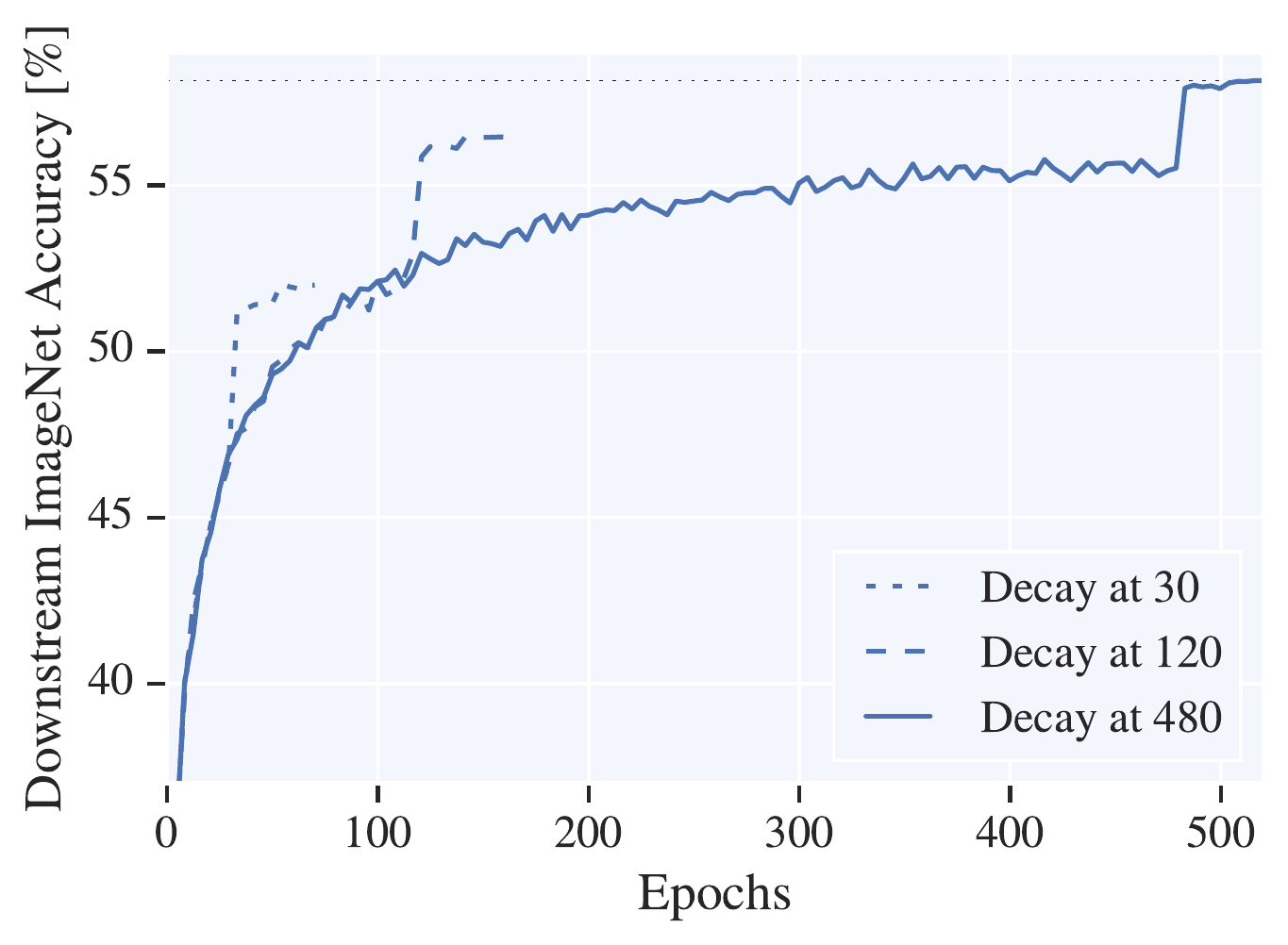}
  \end{center}
  \caption{
     Downstream task accuracy curve of the linear evaluation model trained with SGD on representations from the \emph{Rotation} task.
     The first learning rate decay starts after 30, 120 and 480 epochs.
     We observe that accuracy on the downstream task improves even after very large number of epochs. 
  }\label{fig:downstream_curves_rot}
\end{figure}

Figure~\ref{fig:downstream_curves_rot} depicts how accuracy on our validation split progresses depending on when the learning rate is first decayed.
Surprisingly, we observe that very long training ($\approx 500$ epochs) results in higher accuracy.
Thus, we conclude that SGD optimization hyperparameters play an important role and need to be reported.

\section{Conclusion}\label{sec:conclusion}

In this work, we have investigated self-supervised visual representation learning from the previously unexplored angles. 
Doing so, we uncovered multiple important insights, namely that (1) lessons from architecture design in the fully-supervised setting do not necessarily translate to the self-supervised setting; (2) contrary to previously popular architectures like AlexNet, in residual architectures, the final \emph{pre-logits} layer consistently results in the best performance; (3) the widening factor of CNNs has a drastic effect on performance of self-supervised techniques and (4) SGD training of linear logistic regression may require very long time to converge.
In our study we demonstrated that performance of existing self-supervision techniques can be consistently boosted and that this leads to halving the gap between self-supervision and fully labeled supervision.

Most importantly, though, we reveal that neither is the ranking of architectures consistent across different methods, nor is the ranking  of methods consistent across architectures.
This implies that pretext tasks for self-supervised learning should not be considered in isolation, but in conjunction with underlying architectures.

{\small
\bibliographystyle{ieee}
\bibliography{main}

\begin{thebibliography}{10}\itemsep=-1pt

\bibitem{abadi2016tensorflow}
M.~Abadi, P.~Barham, J.~Chen, Z.~Chen, A.~Davis, J.~Dean, M.~Devin,
  S.~Ghemawat, G.~Irving, M.~Isard, et~al.
\newblock Tensorflow: a system for large-scale machine learning.

\bibitem{behrmann2018invertible}
J.~Behrmann, D.~Duvenaud, and J.-H. Jacobsen.
\newblock Invertible residual networks.
\newblock {\em arXiv preprint arXiv:1811.00995}, 2018.

\bibitem{caron2018deep}
M.~Caron, P.~Bojanowski, A.~Joulin, and M.~Douze.
\newblock Deep clustering for unsupervised learning of visual features.
\newblock {\em European Conference on Computer Vision (ECCV)}, 2018.

\bibitem{chen2018domain}
Y.~Chen, W.~Li, C.~Sakaridis, D.~Dai, and L.~Van~Gool.
\newblock Domain adaptive faster {R-CNN} for object detection in the wild.
\newblock In {\em Conference on Computer Vision and Pattern Recognition
  (CVPR)}, 2018.

\bibitem{dai2018dark}
D.~Dai and L.~Van~Gool.
\newblock Dark model adaptation: Semantic image segmentation from daytime to
  nighttime.
\newblock {\em arXiv preprint arXiv:1810.02575}, 2018.

\bibitem{DinhICLR17}
L.~Dinh, J.~Sohl{-}Dickstein, and S.~Bengio.
\newblock Density estimation using real {NVP}.
\newblock In {\em International Conference on Learning Representations (ICLR)},
  2017.

\bibitem{doersch2015unsupervised}
C.~Doersch, A.~Gupta, and A.~A. Efros.
\newblock Unsupervised visual representation learning by context prediction.
\newblock In {\em International Conference on Computer Vision (ICCV)}, 2015.

\bibitem{doersch2017multi}
C.~Doersch and A.~Zisserman.
\newblock Multi-task self-supervised visual learning.
\newblock In {\em International Conference on Computer Vision (ICCV)}, 2017.

\bibitem{dosovitskiy2014exemplar}
A.~Dosovitskiy, J.~T. Springenberg, M.~Riedmiller, and T.~Brox.
\newblock Discriminative unsupervised feature learning with convolutional
  neural networks.
\newblock In {\em Advances in Neural Information Processing Systems (NIPS)},
  2014.

\bibitem{ebert2018robustness}
F.~Ebert, S.~Dasari, A.~X. Lee, S.~Levine, and C.~Finn.
\newblock Robustness via retrying: Closed-loop robotic manipulation with
  self-supervised learning.
\newblock {\em Conference on Robot Learning (CoRL)}, 2018.

\bibitem{gidaris2018unsupervised}
S.~Gidaris, P.~Singh, and N.~Komodakis.
\newblock Unsupervised representation learning by predicting image rotations.
\newblock In {\em International Conference on Learning Representations (ICLR)},
  2018.

\bibitem{GomezNIPS17}
A.~N. Gomez, M.~Ren, R.~Urtasun, and R.~B. Grosse.
\newblock The reversible residual network: Backpropagation without storing
  activations.
\newblock In {\em Advances in neural information processing systems (NIPS)},
  2017.

\bibitem{goyal2017accurate}
P.~Goyal, P.~Doll{\'a}r, R.~Girshick, P.~Noordhuis, L.~Wesolowski, A.~Kyrola,
  A.~Tulloch, Y.~Jia, and K.~He.
\newblock Accurate, large minibatch sgd: training imagenet in 1 hour.
\newblock {\em arXiv preprint arXiv:1706.02677}, 2017.

\bibitem{he2017mask}
K.~He, G.~Gkioxari, P.~Doll{\'a}r, and R.~Girshick.
\newblock Mask r-cnn.
\newblock In {\em International Conference on Computer Vision (ICCV)}. IEEE,
  2017.

\bibitem{he2015delving}
K.~He, X.~Zhang, S.~Ren, and J.~Sun.
\newblock Delving deep into rectifiers: Surpassing human-level performance on
  imagenet classification.
\newblock In {\em International conference on computer vision (ICCV)}, pages
  1026--1034, 2015.

\bibitem{he2016deep}
K.~He, X.~Zhang, S.~Ren, and J.~Sun.
\newblock Deep residual learning for image recognition.
\newblock In {\em Conference on Computer Vision and Pattern Recognition
  (CVPR)}, 2016.

\bibitem{he2016identity}
K.~He, X.~Zhang, S.~Ren, and J.~Sun.
\newblock Identity mappings in deep residual networks.
\newblock In {\em European conference on computer vision (ECCV)}. Springer,
  2016.

\bibitem{HermansBeyer2017Arxiv}
A.~Hermans, L.~Beyer, and B.~Leibe.
\newblock {In Defense of the Triplet Loss for Person Re-Identification}.
\newblock {\em arXiv preprint arXiv:1703.07737}, 2017.

\bibitem{hinton2012improving}
G.~E. Hinton, N.~Srivastava, A.~Krizhevsky, I.~Sutskever, and R.~R.
  Salakhutdinov.
\newblock Improving neural networks by preventing co-adaptation of feature
  detectors.
\newblock {\em arXiv preprint arXiv:1207.0580}, 2012.

\bibitem{ioffe2015batch}
S.~Ioffe and C.~Szegedy.
\newblock Batch normalization: Accelerating deep network training by reducing
  internal covariate shift.
\newblock {\em International Conference on Machine Learning (ICML)}, 2015.

\bibitem{JacobsenICLR18}
J.~Jacobsen, A.~W.~M. Smeulders, and E.~Oyallon.
\newblock {i-RevNet}: Deep invertible networks.
\newblock In {\em International Conference on Learning Representations (ICLR)},
  2018.

\bibitem{jang2018grasp2vec}
E.~Jang, C.~Devin, V.~Vanhoucke, and S.~Levine.
\newblock {Grasp2Vec}: Learning object representations from self-supervised
  grasping.
\newblock In {\em Conference on Robot Learning}, 2018.

\bibitem{scipy}
E.~Jones, T.~Oliphant, P.~Peterson, et~al.
\newblock {SciPy}: Open source scientific tools for {Python}, 2001.

\bibitem{jouppi2017datacenter}
N.~P. Jouppi, C.~Young, N.~Patil, D.~Patterson, G.~Agrawal, R.~Bajwa, S.~Bates,
  S.~Bhatia, N.~Boden, A.~Borchers, et~al.
\newblock In-datacenter performance analysis of a tensor processing unit.
\newblock In {\em International Symposium on Computer Architecture (ISCA)}.
  IEEE, 2017.

\bibitem{kim2018learning}
D.~Kim, D.~Cho, D.~Yoo, and I.~S. Kweon.
\newblock Learning image representations by completing damaged jigsaw puzzles.
\newblock {\em Winter Conference on Applications of Computer Vision (WACV)},
  2018.

\bibitem{korbar2018cooperative}
B.~Korbar, D.~Tran, and L.~Torresani.
\newblock Cooperative learning of audio and video models from self-supervised
  synchronization.
\newblock {\em arXiv preprint arXiv:1807.00230}, 2018.

\bibitem{KrizhevskyNIPS12}
A.~Krizhevsky, I.~Sutskever, and G.~E. Hinton.
\newblock Imagenet classification with deep convolutional neural networks.
\newblock In {\em Advances in neural information processing systems (NIPS)},
  2012.

\bibitem{krizhevsky2012imagenet}
A.~Krizhevsky, I.~Sutskever, and G.~E. Hinton.
\newblock Imagenet classification with deep convolutional neural networks.
\newblock In {\em Advances in neural information processing systems (NIPS)},
  2012.

\bibitem{lee2018making}
M.~A. Lee, Y.~Zhu, K.~Srinivasan, P.~Shah, S.~Savarese, L.~Fei-Fei, A.~Garg,
  and J.~Bohg.
\newblock Making sense of vision and touch: Self-supervised learning of
  multimodal representations for contact-rich tasks.
\newblock {\em arXiv preprint arXiv:1810.10191}, 2018.

\bibitem{liu1989limited}
D.~C. Liu and J.~Nocedal.
\newblock On the limited memory bfgs method for large scale optimization.
\newblock {\em Mathematical programming}, 45(1-3):503--528, 1989.

\bibitem{mikolov2013efficient}
T.~Mikolov, K.~Chen, G.~Corrado, and J.~Dean.
\newblock Efficient estimation of word representations in vector space.
\newblock {\em arXiv preprint arXiv:1301.3781}, 2013.

\bibitem{mundhenk2018improvements}
T.~N. Mundhenk, D.~Ho, and B.~Y. Chen.
\newblock Improvements to context based self-supervised learning.
\newblock In {\em Conference on Computer Vision and Pattern Recognition
  (CVPR)}, 2018.

\bibitem{nair2010rectified}
V.~Nair and G.~E. Hinton.
\newblock Rectified linear units improve restricted boltzmann machines.
\newblock In {\em International conference on machine learning (ICML)}, 2010.

\bibitem{noroozi2016unsupervised}
M.~Noroozi and P.~Favaro.
\newblock Unsupervised learning of visual representations by solving jigsaw
  puzzles.
\newblock In {\em European Conference on Computer Vision (ECCV)}, 2016.

\bibitem{noroozi2017representation}
M.~Noroozi, H.~Pirsiavash, and P.~Favaro.
\newblock Representation learning by learning to count.
\newblock In {\em International Conference on Computer Vision (ICCV)}, 2017.

\bibitem{noroozi2018boosting}
M.~Noroozi, A.~Vinjimoor, P.~Favaro, and H.~Pirsiavash.
\newblock Boosting self-supervised learning via knowledge transfer.
\newblock {\em Conference on Computer Vision and Pattern Recognition (CVPR)},
  2018.

\bibitem{oord2018representation}
A.~v.~d. Oord, Y.~Li, and O.~Vinyals.
\newblock Representation learning with contrastive predictive coding.
\newblock {\em arXiv preprint arXiv:1807.03748}, 2018.

\bibitem{owens2018audio}
A.~Owens and A.~A. Efros.
\newblock Audio-visual scene analysis with self-supervised multisensory
  features.
\newblock {\em European Conference on Computer Vision (ECCV)}, 2018.

\bibitem{pathak2017learning}
D.~Pathak, R.~B. Girshick, P.~Doll{\'a}r, T.~Darrell, and B.~Hariharan.
\newblock Learning features by watching objects move.
\newblock In {\em Conference on Computer Vision and Pattern Recognition
  (CVPR)}, 2017.

\bibitem{pathakCVPR16context}
D.~Pathak, P.~Kr\"ahenb\"uhl, J.~Donahue, T.~Darrell, and A.~Efros.
\newblock Context encoders: Feature learning by inpainting.
\newblock In {\em Conference on Computer Vision and Pattern Recognition
  ({CVPR})}, 2016.

\bibitem{russakovsky2015imagenet}
O.~Russakovsky, J.~Deng, H.~Su, J.~Krause, S.~Satheesh, S.~Ma, Z.~Huang,
  A.~Karpathy, A.~Khosla, M.~Bernstein, et~al.
\newblock Imagenet large scale visual recognition challenge.
\newblock {\em International Journal of Computer Vision (IJCV)},
  115(3):211--252, 2015.

\bibitem{sayed2018cross}
N.~Sayed, B.~Brattoli, and B.~Ommer.
\newblock Cross and learn: Cross-modal self-supervision.
\newblock {\em arXiv preprint arXiv:1811.03879}, 2018.

\bibitem{schroff2015facenet}
F.~Schroff, D.~Kalenichenko, and J.~Philbin.
\newblock Facenet: A unified embedding for face recognition and clustering.
\newblock In {\em Computer Vision and Pattern Recognition (CVPR)}, 2015.

\bibitem{sermanet2017time}
P.~Sermanet, C.~Lynch, Y.~Chebotar, J.~Hsu, E.~Jang, S.~Schaal, and S.~Levine.
\newblock Time-contrastive networks: Self-supervised learning from video.
\newblock {\em arXiv preprint arXiv:1704.06888}, 2017.

\bibitem{simonyan2014very}
K.~Simonyan and A.~Zisserman.
\newblock Very deep convolutional networks for large-scale image recognition.
\newblock {\em arXiv preprint arXiv:1409.1556}, 2014.

\bibitem{szegedy2015going}
C.~Szegedy, W.~Liu, Y.~Jia, P.~Sermanet, S.~Reed, D.~Anguelov, D.~Erhan,
  V.~Vanhoucke, and A.~Rabinovich.
\newblock Going deeper with convolutions.
\newblock In {\em Conference on Computer Vision and Pattern Recognition
  (CVPR)}, 2015.

\bibitem{wiles2018self}
O.~Wiles, A.~Koepke, and A.~Zisserman.
\newblock Self-supervised learning of a facial attribute embedding from video.
\newblock In {\em British Machine Vision Conference (BMVC)}, 2018.

\bibitem{xie2017aggregated}
S.~Xie, R.~Girshick, P.~Doll{\'a}r, Z.~Tu, and K.~He.
\newblock Aggregated residual transformations for deep neural networks.
\newblock In {\em Conference on Computer Vision and Pattern Recognition
  (CVPR)}. IEEE, 2017.

\bibitem{zagoruyko2016wide}
S.~Zagoruyko and N.~Komodakis.
\newblock Wide residual networks.
\newblock {\em British Machine Vision Conference (BMVC)}, 2016.

\bibitem{zhang2016colorful}
R.~Zhang, P.~Isola, and A.~A. Efros.
\newblock Colorful image colorization.
\newblock In {\em European Conference on Computer Vision (ECCV)}, 2016.

\bibitem{zhang2017split}
R.~Zhang, P.~Isola, and A.~A. Efros.
\newblock Split-brain autoencoders: Unsupervised learning by cross-channel
  prediction.
\newblock In {\em Conference on Computer Vision and Pattern Recognition
  (CVPR)}, 2017.

\end{thebibliography}
}

\clearpage
\appendix

\begin{figure*}[t]
  \begin{center}
    \includegraphics[width=0.9\linewidth]{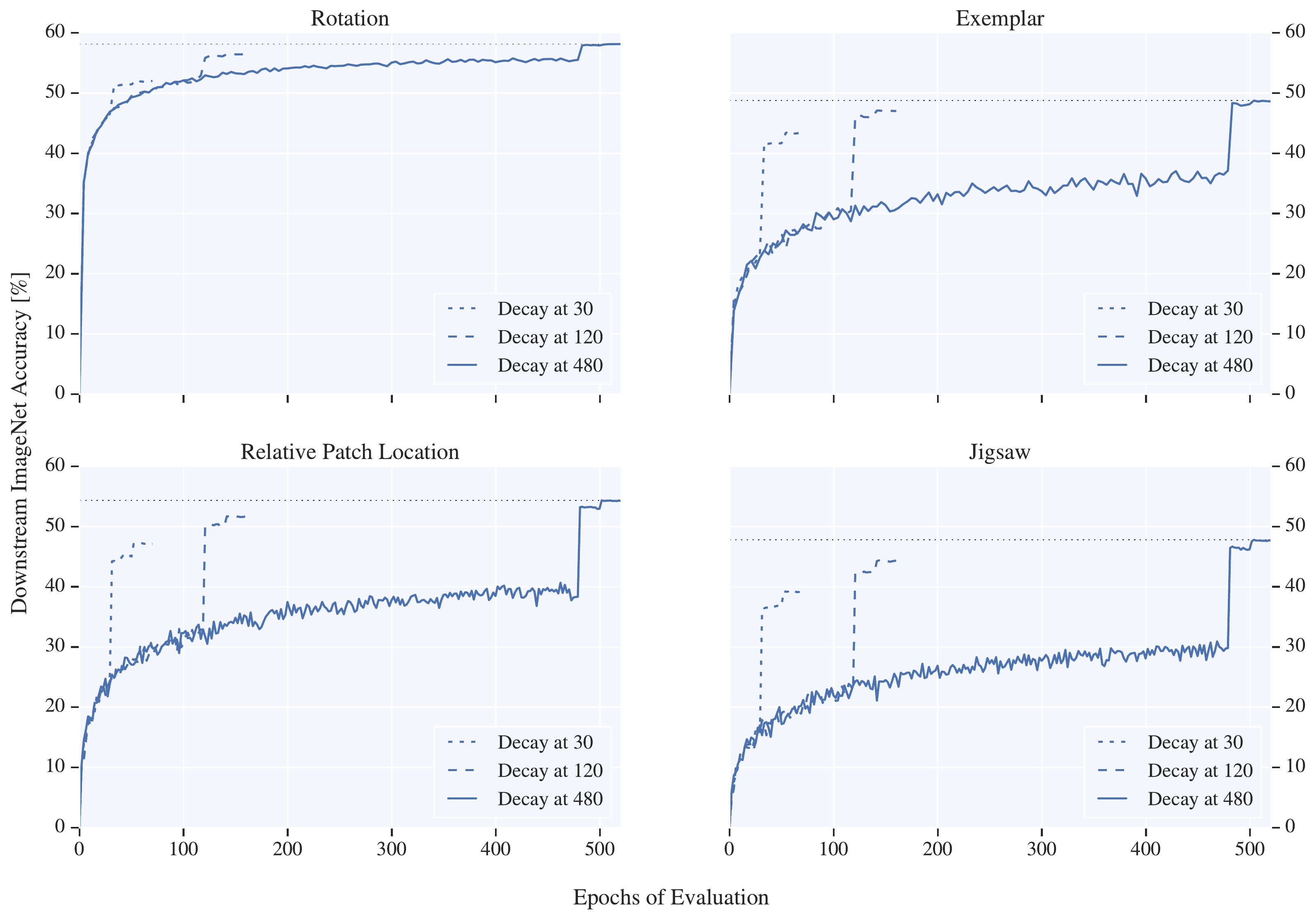}
  \end{center}
  \caption{
     Downstream task accuracy curve of the linear evaluation model trained with SGD on representations learned by the four self-supervision pretext tasks.
  }\label{fig:downstream_curves}
\end{figure*}

\section{Self-supervised model details}

For training all self-supervised models we use stochastic gradient descent (SGD) with momentum.
The initial learning rate is set to $0.1$ and the momentum coefficient is set to $0.9$.
We train for 35 epochs in total and decay the learning rate by a factor of 10 after 15 and 25 epochs.
As we use large mini-batch sizes $B$ during training, we leverage two recommendations from~\cite{goyal2017accurate}:
(1) a learning rate scaling, where the learning rate is multiplied by
$\frac{B}{256}$
and (2) a linear learning rate warm-up during the initial 5 epochs.

In the following we give additional details that are specific to the choice of self-supervised learning technique.

\PAR{Rotation:} During training we use the data augmentation mechanism from~\cite{szegedy2015going}. 
We use mini-batches of $B=1024$ images, where each image is repeated 4 times: once for every rotation. 
The model is trained on 128 TPU~\cite{jouppi2017datacenter} cores.

\PAR{Exemplar:} In order to generate image examples, we use the data augmentation mechanism from~\cite{szegedy2015going}.
During training, we use mini-batches of size $B=512$, and
for each image in a mini-batch we randomly generate 8 examples.
We use an implementation\footnote{\url{https://www.tensorflow.org/api_docs/python/tf/contrib/losses/metric_learning/triplet_semihard_loss}} of the triplet loss~\cite{schroff2015facenet} from the \texttt{tensorflow} package~\cite{abadi2016tensorflow}.
The margin parameter of the triplet loss is set to $0.5$.
We use 32 TPU cores for training.

\PAR{Jigsaw:} Similar to~\cite{noroozi2016unsupervised}, we preprocess the input images by:
(1) resizing the input image to $292 \times 292$ and randomly cropping it to $255 \times 255$;
(2) converting the image to grayscale with probability \sfrac{2}{3} by averaging the color channels;
(3) splitting the image into a $3 \times 3$ regular grid of cells (size $85 \times 85$ each) and randomly cropping $64 \times 64$-sized patches inside every cell; 
(4) standardize every patch individually such that its pixel intensities have zero mean and unit variance.
We use SGD with batch size $B=1024$.
For each image individually, we randomly select 16 out of the 100 pre-defined permutations and apply all of them.
The model is trained on 32 TPU cores.

\PAR{Relative Patch Location:} We use the same patch prepossessing, representation extraction and training setup as in the Jigsaw model. The only difference is the loss function as discussed in the main text.

\section{Downstream training details}

\PAR{Training linear models with SGD:}
For training linear models with SGD, we use a standard data augmentation technique in the \emph{Rotation} and \emph{Exemplar} cases:
(1) resize the image, preserving its aspect ratio such that its smallest side is 256.
(2) apply a random crop of size $224 \times 224$.
For the patch-based methods, we extract representations by averaging the representations of all nine, colorful, standardized patches of an image.
At final evaluation-time, fixed patches are obtained by scaling the image to $255 \times 255$, cropping the central $192 \times 192$ patch and taking the $3 \times 3$ grid of $64 \times 64$-sized patches from it.

We use a batch-size of $2048$ for evaluation of representations from \emph{Rotation} and \emph{Exemplar} models and of $1024$ for \emph{Jigsaw} and \emph{Relative Patch Location} models.
As we use large mini-batch sizes, we perform learning-rate scaling, as suggested in~\cite{goyal2017accurate}.

\PAR{Training linear models with L-BFGS:}
We use a publicly available implementation of the L-BFGS algorithm~\cite{liu1989limited} from the \emph{scipy}~\cite{scipy} package with the default parameters and set the maximum number of updates to 800.
For training all our models we apply $l_2$ penalty $\lambda ||W||^2_2$, where 
${W \in \mathbb{R}^{M \times C}}$ is the matrix of model parameters, $M$ is the size of the representation, and $C$ is the number of classes.
We set $\lambda = \frac{100.0}{MC}$.

\PAR{Training MLP models with SGD:} In the MLP evaluation scenario, we use a single hidden layer with $1000$ channels.
At training time, we apply dropout~\cite{hinton2012improving} to the hidden layer with a drop rate of 50\%.
The $l_2$ regularization scheme is the same as in the \emph{L-BFGS} setting.
We optimize the MLP model using stochastic gradient descent with momentum (the momentum coefficient is 0.9) for 180 epochs.
The batch size is 512, initial learning rate is 0.01 and we decay it twice by a factor of 10: after 60 and 120 epochs.

%------------------------------------------------------------------------
\section{Training linear models with SGD}

In Figure~\ref{fig:downstream_curves} we demonstrate how accuracy on the validation data progresses
during the course of SGD optimization.
We observe that in all cases achieving top accuracy requires training for a very large number of epochs.

%------------------------------------------------------------------------
\section{More Results on Places205 and ImageNet}

For completeness, we present full result tables for various settings considered in the main paper. 
These include numbers for ImageNet evaluated on \SI{10}{\percent} of the data (Table~\ref{tbl:bigtable_imagenet_10}) as well as all results when evaluating on the Places205 dataset (Table~\ref{tbl:bigtable_places}) and a random subset of \SI{5}{\percent} of the Places205 dataset (Table~\ref{tbl:bigtable_places_5}).

Finally, Table~\ref{tbl:sota-full} is an extended version of Table~\ref{tbl:sota} in the main paper, additionally providing the top-5 accuracies of our various best models on the public ImageNet validation set.

\begin{table*}[h]
  \caption{Evaluation on ImageNet with \SI{10}{\percent} of the data.}
  \label{tbl:bigtable_imagenet_10}
  \setlength{\tabcolsep}{0pt}
  \setlength{\extrarowheight}{5pt}
  \renewcommand{\arraystretch}{0.75}
  \centering
  \begin{tabularx}{\linewidth}{p{2.7cm}CCCCp{7pt}CCCp{7pt}CCp{7pt}CC}
    \toprule[1pt]
    \multirow{3}{*}{Model} & \multicolumn{4}{c}{Rotation} && \multicolumn{3}{c}{Exemplar} && \multicolumn{2}{c}{RelPatchLoc} && \multicolumn{2}{c}{Jigsaw}\\
    \cmidrule[0.5pt]{2-5} \cmidrule[0.5pt]{7-9} \cmidrule[0.5pt]{11-12} \cmidrule[0.5pt]{14-15}
     & $4\times$ & $8\times$ & $12\times$ & $16\times$ && $4\times$ & $8\times$ & $12\times$ && $4\times$ & $8\times$ && $4\times$ & $8\times$\\

    \midrule
    
    RevNet50 & 31.3 & 34.6 & 37.9 & 38.4 && 27.1 & 30.0 & 31.1 &&  24.6 & 27.8 && 25.0 & 24.2 \\
    ResNet50 v2 & 28.2 & 31.7 & 32.2 & 33.3 && 28.3 & 30.1 & 31.2 && 25.8 & 29.4 && 23.3 & 24.1 \\
    ResNet50 v1 & 26.8 & 27.2 & 27.4 & 27.8 && 28.7 & 30.8 & 31.7 && 30.2 & 33.2 && 26.4 & 28.3 \\

    \arrayrulecolor{lightgray}\midrule[0.25pt]\arrayrulecolor{black}
    RevNet50 (-) & 30.2 & 32.3 & 33.3 & 33.4 && 25.7 & 26.3 & 26.4 && 21.6 & 25.0 && 24.1 & 24.9 \\
    ResNet50 v2 (-) & 28.4 & 28.6 & 28.2 & 28.5 && 26.5 & 27.3 & 27.3 && 26.1 & 26.3 && 23.9 & 23.1 \\

    \arrayrulecolor{lightgray}\midrule[0.25pt]\arrayrulecolor{black}
    VGG19-BN & \leavevmode\hphantom{0}8.8 & \leavevmode\hphantom{0}6.7 & \leavevmode\hphantom{0}7.6 & 13.1 && 16.6 & 17.7 & 18.2 && 15.8 & 16.8 && 10.6 & 10.7 \\

    \bottomrule
  \end{tabularx}
\end{table*}

\clearpage

\begin{table*}[h]
  \caption{Evaluation on Places205.}
  \label{tbl:bigtable_places}
  \setlength{\tabcolsep}{0pt}
  \setlength{\extrarowheight}{5pt}
  \renewcommand{\arraystretch}{0.75}
  \centering
  \begin{tabularx}{\linewidth}{p{2.7cm}CCCCp{7pt}CCCp{7pt}CCp{7pt}CC}
    \toprule[1pt]
    \multirow{3}{*}{Model} & \multicolumn{4}{c}{Rotation} && \multicolumn{3}{c}{Exemplar} && \multicolumn{2}{c}{RelPatchLoc} && \multicolumn{2}{c}{Jigsaw}\\
    \cmidrule[0.5pt]{2-5} \cmidrule[0.5pt]{7-9} \cmidrule[0.5pt]{11-12} \cmidrule[0.5pt]{14-15}
     & $4\times$ & $8\times$ & $12\times$ & $16\times$ && $4\times$ & $8\times$ & $12\times$ && $4\times$ & $8\times$ && $4\times$ & $8\times$\\

    \midrule

    RevNet50 & 41.8 & 45.3 & 47.4 & 47.9 && 39.4 & 43.1 & 44.5 && 37.5 & 41.9 && 37.1 & 40.7 \\
    ResNet50 v2 & 39.8 & 43.2 & 44.2 & 44.8 && 39.5 & 42.8 & 44.3 && 38.7 & 43.2 && 36.3 & 39.2 \\
    ResNet50 v1 & 38.1 & 40.0 & 41.3 & 42.0 && 39.3 & 43.1 & 44.5 && 42.3 & 46.2 && 39.4 & 42.9 \\
    \arrayrulecolor{lightgray}\midrule[0.25pt]\arrayrulecolor{black}
    RevNet50 (-) & 39.5 & 44.3 & 46.3 & 47.5 && 35.8 & 39.3 & 40.7 && 32.5 & 39.7 && 34.5 & 38.5 \\
    ResNet50 v2 (-) & 35.5 & 39.5 & 41.8 & 42.8 && 32.6 & 34.9 & 36.0 && 35.8 & 39.1 && 31.6 & 33.2 \\
    \arrayrulecolor{lightgray}\midrule[0.25pt]\arrayrulecolor{black}
    VGG19-BN & 22.6 & 21.6 & 23.8 & 30.7 && 29.3 & 32.0 & 33.3 && 31.5 & 33.6 && 24.6 & 27.2 \\

    \bottomrule
  \end{tabularx}
\end{table*}

\begin{table*}[b]
  \caption{Evaluation on Places205 with \SI{5}{\percent} of the data.}
  \label{tbl:bigtable_places_5}
  \setlength{\tabcolsep}{0pt}
  \setlength{\extrarowheight}{5pt}
  \renewcommand{\arraystretch}{0.75}
  \centering
  \begin{tabularx}{\linewidth}{p{2.7cm}CCCCp{7pt}CCCp{7pt}CCp{7pt}CC}
    \toprule[1pt]
    \multirow{3}{*}{Model} & \multicolumn{4}{c}{Rotation} && \multicolumn{3}{c}{Exemplar} && \multicolumn{2}{c}{RelPatchLoc} && \multicolumn{2}{c}{Jigsaw}\\
    \cmidrule[0.5pt]{2-5} \cmidrule[0.5pt]{7-9} \cmidrule[0.5pt]{11-12} \cmidrule[0.5pt]{14-15}
     & $4\times$ & $8\times$ & $12\times$ & $16\times$ && $4\times$ & $8\times$ & $12\times$ && $4\times$ & $8\times$ && $4\times$ & $8\times$\\

    \midrule

    RevNet50 & 32.1 & 33.4 & 34.5 & 34.8     && 30.7 & 31.2 & 31.6 && 28.9 & 29.7 && 29.3 & 29.3 \\
    ResNet50 v2 & 30.6 & 31.8 & 31.8 & 32.0  && 32.1 & 31.8 & 32.2 && 29.8 & 31.1 && 29.4 & 28.9 \\
    ResNet50 v1 & 30.0 & 29.2 & 29.0 & 29.2  && 32.5 & 32.5 & 32.7 && 33.2 & 33.9 && 31.2 & 31.3 \\
    \arrayrulecolor{lightgray}\midrule[0.25pt]\arrayrulecolor{black}
    RevNet50 (-) & 33.5 & 34.4 & 34.5 & 34.3    && 31.0 & 32.2 & 32.2 && 27.4 & 30.8 && 29.8 & 31.1 \\
    ResNet50 v2 (-) & 31.6 & 33.2 & 33.6 & 33.6 && 30.0 & 31.4 & 31.9 && 30.9 & 31.9 && 28.4 & 28.9 \\
    \arrayrulecolor{lightgray}\midrule[0.25pt]\arrayrulecolor{black}
    
    VGG19-BN & 16.8 & 13.9 & 15.3 & 20.2 && 23.5 & 23.4 & 23.7 && 23.8 & 24.0 && 19.3 & 18.7 \\

    \bottomrule
  \end{tabularx}
\end{table*}

\begin{table*}[b]
  \setlength{\tabcolsep}{0pt}
  \setlength{\extrarowheight}{5pt}
  \renewcommand{\arraystretch}{0.75}
  \centering
  \caption{
    Comparison of the published self-supervised models to our best models.
    The scores correspond to accuracy of linear logistic regression that is trained on top of representations provided by self-supervised models.
    Official validation splits of \emph{ImageNet} and \emph{Places205} are used for computing accuracies.
    The ``Family'' column shows which basic model architecture was used in the referenced literature: \textbf{A}lexNet, \textbf{V}GG-style, or \textbf{R}esidual.}  \label{tbl:sota-full}
  \begin{tabularx}{\linewidth}{p{0.4cm}p{1pt}p{3.7cm}CCCp{7pt}CCC}
    \toprule[1pt]
     \multirow{3}{*}{\rotatebox{90}{\hspace*{5pt}Family}} && & \multicolumn{3}{c}{ImageNet} && \multicolumn{3}{c}{Places205}\\
    \cmidrule[0.5pt]{4-6} \cmidrule[0.5pt]{8-10}
     && & Prev. top1 & Ours top1 & Ours top5 && Prev. top1 & Ours top1 & Ours top5\\
    \midrule

    A && Rotation\cite{gidaris2018unsupervised} & 38.7 & \textbf{55.4} & \textbf{77.9} && 35.1 & \textbf{48.0} & \textbf{77.9} \\
    R && Exemplar\cite{doersch2017multi} & 31.5 & 46.0 & 68.8 && - & 42.7 & 72.5 \\
    R && Rel. Patch Loc.\cite{doersch2017multi} & 36.2 & 51.4 & 74.0 && - & 45.3 & 75.6 \\
    A && Jigsaw\cite{noroozi2016unsupervised,zhang2017split} & 34.7 & 44.6 & 68.0 && 35.5 & 42.2 & 71.6 \\

    \arrayrulecolor{lightgray}\midrule[0.25pt]\arrayrulecolor{black}

    R && Supervised RevNet50    & 74.8 & 74.4 & 91.9 &&   -  & 58.9 & 87.5 \\
    R && Supervised ResNet50 v2 & 76.0 & 75.8 & 92.8 &&   -  & 61.6 & 89.0 \\
    V && Supervised VGG19       & 72.7 & 75.0 & 92.3 && 58.9 & 61.5 & 89.3 \\
    \bottomrule
  \end{tabularx}
\end{table*}

\end{document}